\documentclass[journal,compsoc]{IEEEtran}

\ifCLASSOPTIONcompsoc
  \usepackage[nocompress]{cite}
\else
  \usepackage{cite}
\fi

\ifCLASSINFOpdf
\else
\fi

\usepackage{lipsum}
\usepackage{rotating}
\usepackage{multirow}
\usepackage[table,xcdraw]{xcolor}
\usepackage{algorithm,algpseudocode}
\usepackage{amsmath}
\usepackage{amsfonts}
\usepackage{amssymb}

\newcommand\StateX{\Statex\hspace{\algorithmicindent}}

\begin{document}
%
\title{Rethinking Recurrent Neural Networks and Other Improvements for Image Classification}

\author{Nguyen~Huu~Phong,
        ~Bernardete~Ribeiro
\IEEEcompsocitemizethanks{\IEEEcompsocthanksitem Nguyen Huu Phong is with Department of Informatics Engineering (DEI), Center for Informatics and Systems of the University of Coimbra (CISUC), University of Coimbra, Coimbra, Portugal\\
E-mail: phong@dei.uc.pt
\IEEEcompsocthanksitem Bernardete Ribeiro is with Department of Informatics Engineering (DEI), Center for Informatics and Systems of the University of Coimbra
(CISUC), University of Coimbra, Coimbra, Portugal\\
E-mail: bribeiro@dei.uc.pt}
}

\maketitle

\begin{abstract}
Over the long history of machine learning, which dates back several decades, recurrent neural networks (RNNs) have been used mainly for sequential data and time series and generally with 1D information. Even in some rare studies on 2D images, these networks are used merely to learn and generate data sequentially rather than for image recognition tasks. In this study, we propose integrating an RNN as an additional layer when designing image recognition models. We also develop end-to-end multimodel ensembles that produce expert predictions using several models. In addition, we extend the training strategy so that our model performs comparably to leading models and can even match the state-of-the-art models on several challenging datasets (e.g., SVHN (0.99), Cifar-100 (0.9027) and Cifar-10 (0.9852)). Moreover, our model sets a new record on the Surrey dataset (0.949). The source code of the methods provided in this article is available at https://github.com/leonlha/e2e-3m and http://nguyenhuuphong.me.
\end{abstract}


%
\IEEEpeerreviewmaketitle

\section{Introduction}
Recently, the image recognition task has been transformed by the availability of high-performance computing hardware---particularly modern graphical processing units (GPUs) and large-scale datasets. The early designs of convolutional neural networks (ConvNets) in the 1990s were shallow and included only a few layers; however, as the ever-increasing volume of image data with higher resolutions required a concomitant increase in computing power, the field has evolved; modern ConvNets have deeper and wider layers with improved efficiency and accuracy~\cite{simonyan2014very,szegedy2016rethinking,szegedy2017inception,he2016deep,xie2017aggregated,chollet2017xception,hu2018squeeze}. The later developments involved a balancing act among network depth, width and image resolution~\cite{tan2019efficientnet} and determining appropriate augmentation policies~\cite{cubuk2019autoaugment}.

During this same period, recurrent neural networks have proven successful at various applications, including natural language processing~\cite{liu2016recurrent,yin2017comparative}, machine translation~\cite{chopra2016abstractive}, speech recognition~\cite{chiu2018state,chan2016listen}, weather forecasting~\cite{qing2018hourly}, human action recognition~\cite{zhang2017view,song2017end,liu2017enhanced}, drug discovery~\cite{segler2018generating}, and so on. However, in the image recognition field, RNNs are largely used merely to generate image pixel sequences~\cite{gregor2015draw,oord2016pixel} rather than being applied for whole-image recognition purposes.

As the architecture of RNNs has evolved through several forms and become optimized, it could be interesting to study whether these spectacular advances have a direct effect on image classification. In this study, we take a distinct approach by integrating an RNN and considering it as an essential layer when designing an image recognition model.

In addition, we propose end-to-end (E2E) multiple-model ensembles that learn expertise through the various models. This approach was the result of a critical observation: when training models for specific datasets, we often select the most accurate models or group some models into an ensemble. We argue that merged predictions can provide a better solution than can a single model. Moreover, the ensembling process essentially breaks the complete operation (from obtaining input data to final prediction) into separate stages and each step may be performed on different platforms. This approach can cause serious issues that may even make it impossible to integrate the operation into a single location (e.g., on real-time systems)~\cite{wang2019real,shi2016real} or to a future platform such as a system on a chip (SoC)~\cite{giri2020esp4ml,zhu2018mobile}.

Additionally, we explore other key techniques, such as the learning rate strategy and test-time augmentation, to obtain overall improvements. The remainder of this article is organized as follows. Our main contributions are discussed in Section \ref{contributions}. Various RNN formulations, including a typical RNN and more advanced RNNs, are the focus of subsection \ref{rnn}. In addition, our core idea for designing ConvNet models that gain experience from expert models is highlighted in subsection \ref{e2e}. In subsection \ref{experiments1}, we evaluate our design on the iNaturalist dataset, and in subsection \ref{experiments2}, the performances of various image recognition models integrated with RNNs are thoroughly analyzed. In subsection \ref{experiments3}, we evaluate our design on the iCassava dataset. In subsections \ref{experiments4} and \ref{experiments5}, we discuss our learning rate strategy and the softmax pruning technology. Additionally, we extend our experiments in subsection \ref{extension} to include more challenging datasets (i.e., SVHN, Cifar-100, and Cifar-10) and show that our model's performance can match the state-of-the-art models even under limited resources. In subsection \ref{stateoftheart}, we show that our approach outperforms the state-of-the-art methods by a large margin. Finally, we conclude our research in Section \ref{conclusions}.

\section{Contributions}
\label{contributions}
Our research differs from previous works in several ways. First, most studies utilize RNNs for sequential data and time series. Except for rare cases, RNNs are used largely to generate sequences of image pixels. Instead, we propose integrating RNNs as an essential layer of ConvNets.

Second, we present our core idea for designing a ConvNet in which the model is able to learn decisions from expert models. Typically, we choose predictions from a single model or an ensemble of models.

Another main contribution of this work involves a training strategy that allows our models to perform competitively, matching previous approaches on several datasets and outperforming some state-of-the-art models.

We make our source code so other researchers can replicate this work. The program is written in the Jupyter Notebook environment using a web-based interface with a few extra libraries.

\section{Methodologies}
Our key idea is to integrate an RNN layer into ConvNet models. We propose several RNNs and present the computational formulas. The concept of training a model to learn predictions from multiple individual models and the design of such a model is also discussed.
\label{methodologies}

\subsection{Recurrent Neural Networks}
\label{rnn}
For the purpose of performance analysis and comparison, we adopt both a typical RNN and some more advanced RNNs i.e., . (a long short term memory (LSTM) and gated recurrent unit (GRU) as well as a bidirectional RNN (BRNN)). The formulations of the selected RNNs are presented as follows.

Considering a standard RNN with a given input sequence ${x_1, x_2,...,x_T}$, the hidden cell state is updated at a time step $t$ as follows:
\begin{equation}
h_t=\sigma(W_h h_{t-1}+W_x x_t+b),
\end{equation}
where $W_h$ and $W_x$ denote weight matrices, b represents the bias, and $\sigma$ is a sigmoid function that outputs values between 0 and 1.

The output of a cell, for ease of notation, is defined as 
\begin{equation}
y_t=h_t,
\end{equation}
but can also be shown using the $softmax$ function, in which $\hat{y}_t$ is the output and $y_t$ is the target:
\begin{equation}
\hat{y_t}=softmax(W_y h_t+b_y).
\end{equation}
A more sophisticated RNN or LSTM that includes the concept of a forget gate can be expressed as shown in the following equations:
\begin{gather}
f_t=\sigma(W_{fh} h_{t-1}+W_{fx} x_t+b_f),\\
i_t=\sigma(W_{ih} h_{t-1}+W_{ix} x_t+b_i),\\
c'_t=tanh(W_{c'h} h_{t-1}+W_{c'x} x_t+b'_c),\\
c_t=f_t \odot c_{t-1}+i_t \odot c'_t,\\
o_t=\sigma(W_{oh} h_{t-1}+W_{ox} x_t+b_o),\\
h_t=o_t \odot tanh(c_t),
\end{gather}
where the $\odot$ operation represents an elementwise vector product, and $f$, $i$, $o$ and $c$ are the forget gate, input gate, output gate and cell state, respectively. Information is retained when the forget gate $f_t$ becomes 1 and eliminated when $f_t$ is set to 0.

Because LSTMs require powerful computing resources, we use a variation, (i.e., a GRU) for optimization purposes. The GRU combines the input gate and forget gate into a single gate---namely, the update gate. The mathematical formulas are expressed as follows:
\begin{gather}
r_t=\sigma(W_{rh} h_{t-1}+W_{rx} x_t+b_r),
\\
z_t=\sigma(W_{zh} h_{t-1}+W_{zx} x_t+b_z),
\\
h'_t=tanh(W_{h'h} (r_t \odot h_{t-1})+W_{h'x} x_t+b_z),
\\
h_t=(1-z_t) \odot h_{t-1}+z_t \odot h'_t.
\end{gather}

Finally, while a typical RNN essentially takes only previous information, bidirectional RNNs integrate both past and future information:
\begin{gather}
h_t=\sigma(W_{hx} x_t+W_{hh} h_{t-1} +b_h),
\\
z_t=\sigma(W_{ZX} x_t+W_{HX} h_{t+1} +b_z),
\\
\hat{y_t}=softmax(W_{yh} h_t+W_{yz} z_t+b_y),
\end{gather}
where $h_{t-1}$ and $h_{t+1}$ indicate hidden cell states at the previous time step ($t-1$) and the future time step ($t+1$).

For more details on the RNN, LSTM, GRU and BRNN models, please refer to the following articles ~\cite{jing2019gated,salehinejad2017recent,yu2019review}, ~\cite{graves2013generating,gers1999learning},
~\cite{cho2014learning,jing2019gated} and ~\cite{lipton2015critical,schuster1997bidirectional} respectively.

\subsection{End-to-end Ensembles of Multiple Models}
\label{e2e}
\begin{figure}[ht!]
\begin{center}
\includegraphics[width=0.5\textwidth]{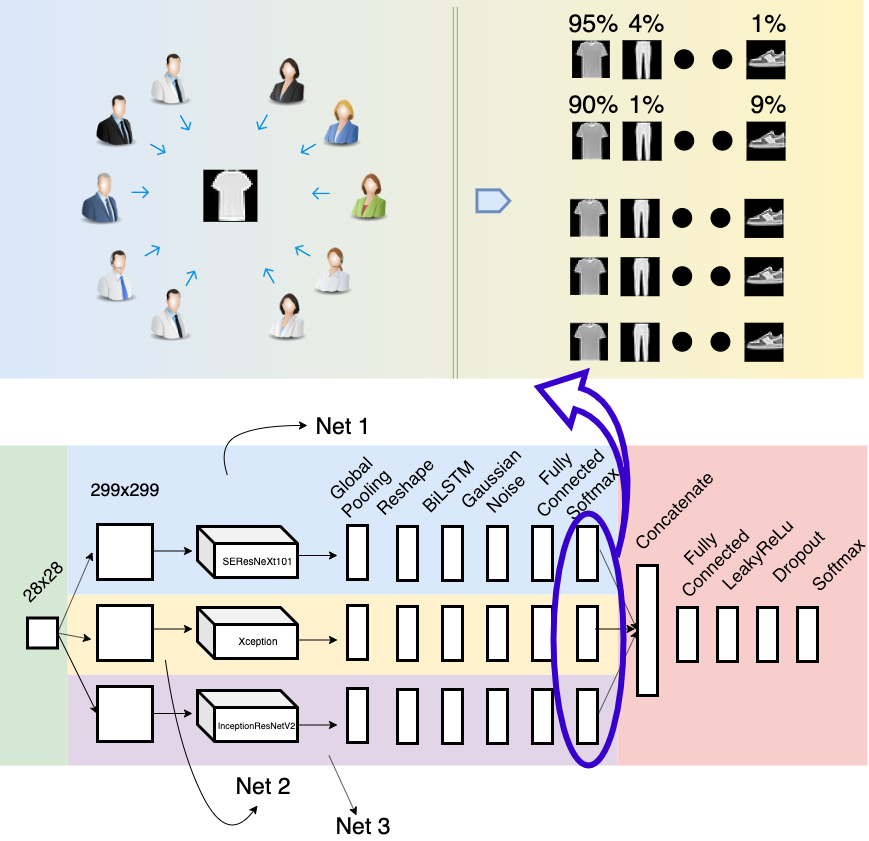}
\caption{End-to-end Ensembles of Multiple Models: Concept and Design. The upper part of this image illustrates our key idea, in which several actors make predictions for a sample and output the probability of the sample belonging to each category. The lower part presents a ConvNet design in which three distinct and recent models are aggregated and trained using advanced neural networks. This design also illustrates our proposal of integrating an RNN into an image recognition model. (Best viewed in color)}
\label{fig:mul_models}
\end{center}
\end{figure}
Our main idea for this design is that when several models are trained on a certain dataset, we would typically choose the model that yields the best accuracy. However, we could also construct a model that combines expertise from all the individual models. We illustrate this idea in Figure \ref{fig:mul_models}.

As shown in the upper part, each actor represents a trained single model. For each presented sample, each actor predicts the probability that the sample belongs to each category. These probabilities are then combined and utilized to train our model.

The bottom part of the figure represents our ConvNet design for this idea. We essentially select three models to make predictions instead of two. Using only two models may result in a situation where one model dominates the other. In other words, we would use the predictions primarily from only one model. Adding one additional model provides a balance that offsets this weakness of a two-model architecture. Note that we limit our designs to just three models because of resource limitations (i.e., GPU memory). We name this model E2E-3M, where ``E2E`` is an abbreviation of the ``end-to-end`` learning process~\cite{he2019streaming,bojarski2017explaining,bojarski2016end}, in which a model performs all phases from training until final prediction. The `` 3M`` simply documents the combination of three models.

Each individual model (named Net 1, Net 2, and Net 3) is a fine-tuned model in which the last layer is removed and replaced by more additional layers, e.g., global pooling to reduce the network size and an RNN module (including a reshape layer and an RNN layer). The models also employ Gaussian noise to prevent overfitting, include a fully connected layer and each model has its own softmax layer. The outputs from these three models are concatenated and utilized to train the subsequent neural network module, which consists of a fully connected layer, a LeakyReLU\cite{maas2013rectifier} layer, a dropout\cite{srivastava2014dropout} layer and a softmax layer for classification.

Ensemble learning is a key aspect of this design. An ensemble refers to an aggregation of weaker models (base learners) combined to construct a model that performs more efficiently (a strong learner)~\cite{sagi2018ensemble}. The ensembling technique is more prevalent in machine learning than in deep learning, especially in the image recognition field, because convolution requires considerable computational power. Most of the recent related studies focus on a simple averaging or voting mechanism~\cite{perez2019solo,antipov2016minimalistic,duan2017ensemble,vasan2020image,wang2019blur}, and few have investigated integrating trainable neural networks~\cite{amin2019ensemble,fan2018multi}. Our research differs from prior studies, because we study the design on a much larger scale using a larger number of up-to-date ConvNets.

Suppose that our ConvNet design has $n$ fine-tuned models or classifiers and that the dataset contains $c$ classes. The output of each classifier can be represented as a distribution vector:
\begin{equation}
\Delta_j=[\delta_{1j}\quad\delta_{2j}\quad\dots\quad\delta_{cj}],
\end{equation}
where
\begin{gather*}
1 \leq j \leq n,
\\
0 \leq \delta_{ij} \leq 1 \quad \forall 1 \leq i \leq c,
\\
\sum\limits_{i=1}^c \delta_{ij} = 1.
\end{gather*}

After concatenating the outputs of the $n$ classifiers, the distribution vector becomes 
\begin{equation}
\Delta=[\Delta_1 \quad\Delta_2\quad\dots\quad\Delta_n].
\end{equation}

For formulaic convenience, we assume that the neural network has only one layer and that the number of neurons is equal to the number of classes. As usual, the networks' weights, $\Theta$, are initialized randomly. The distribution vector is computed as follows:
\begin{equation}
\Delta^{'}=\Theta \cdot \Delta=[\delta_1^{'}\quad\delta_2^{'}\quad\dots\quad\delta_c^{'}],
\end{equation}
where
\begin{gather*}
1 \leq j \leq c,
\\
\delta_j^{'} = \sum\limits_{i=1}^{nc} \delta_{i} w_{ij}.
\end{gather*}

Finally, after the softmax activation, the output is 
\begin{equation}
\eta_j^{'}=\frac{e^{\delta_j^{'}}}{\sum\limits_{i=1}^c e^{\delta_j^{'}}}.
\end{equation}

\section{Experiments}
\label{experiments}
In this section, we will present the experiments used to evaluate our initial design (without an RNN) and then analyze the performance when RNNs are integrated. We also describe our end-to-end ensemble multiple models, the developed training strategy, and the extension of the softmax layer. These experiments were performed on the iNaturalist'19~\cite{van2018inaturalist} and iCassava'19 Challenges~\cite{mwebaze2019icassava}, and on the Cifar-10~\cite{krizhevsky2009learning} and Fashion-MNIST~\cite{fashion_mnist} datasets. We also extend our experiments on Cifar-100~\cite{krizhevsky2009learning}, SVHN~\cite{netzer2011reading} and Surrey~\cite{pugeault2011spelling}.

\subsection{Experiment 1}
\label{experiments1}
Deep learning~\cite{lecun2015deep} and convolutional neural networks (ConvNets) have achieved notable successes in the image recognition field. From the early LeNet model~\cite{lecun1989backpropagation}, first proposed several decades ago, to the recent AlexNet ~\cite{krizhevsky2012imagenet}, Inception~\cite{szegedy2015going,szegedy2016rethinking,szegedy2017inception}, ResNet~\cite{he2016deep}, SENet~\cite{hu2018squeeze} and EfficientNet~\cite{tan2019efficientnet} models, these ConvNets have leveraged automated classification to exceed human performance in several applications. This efficiency is due to the high availability of powerful computer hardware, specifically GPUs and big data.

In this subsection, we examine the significance of our design, which utilizes several ConvNets constructed based on leading architectures such as InceptionV3, ResNet50, Inception-ResNetV2, Xception, MobileNetV1 and SEResNeXt101. We adopt InceptionV3 as a baseline model since the popularity of ConvNets among deep learning researchers means they can often be used as standard testbeds. In addition, InceptionV3 is known for employing sliding kernels, e.g., $1 \times $1, $3 \times $3 or $5 \times $5, in parallel, which essentially reduces the computation and increases the accuracy. We also use the simplest version of residual networks, i.e., . ResNet50 (ResNet has several versions, including ResNet50, ResNet101 and ResNet152, named according to the number of depth layers), which introduces short circuits through each network layer that greatly reduce the training time. In addition, we explore other ConvNets to facilitate the comparisons and evaluations.

\begin{figure}[!ht]
\begin{center}
\includegraphics[width=0.5\textwidth]{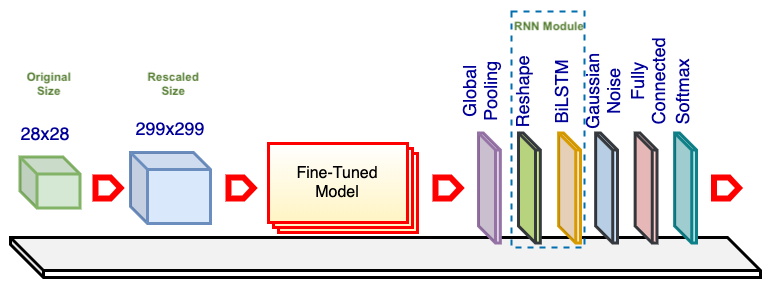}
\caption{Single E2E-3M Model. The fine-tuned model is a pretrained ConvNet (e.g., InceptionV3) with the top layers excluded and the weights retrained. The base model comes preloaded with ImageNet weights. The original images are rescaled as needed to match the required input size of the fine-tuned model or for image resolution analysis. The global pooling layer reduces the networks' size, and the reshaping layer converts the data to a standard input form for the RNN layer. The Gaussian noise layer improves the variation among samples to prevent overfitting. The fully connected layer aims to improve classification. The softmax layer is another fully connected layer that has the same number of neurons as the number of dataset categories, and it utilizes softmax activation.}
\label{fig:sgl_model}
\end{center}
\end{figure}

Our initial design is depicted in Figure~\ref{fig:sgl_model}. The core of this process implements one of the abovementioned ConvNet models. While the architectures of these ConvNets differ, usually their top layers function as classifiers and can be replaced to adapt the models to different datasets. For example, Xception and ResNet50 use a global pooling and a fully connected layer as the top layers, while VGG19~\cite{zeiler2014visualizing} uses a flatten layer and two fully connected layers (the original article used max pooling, but for some reason the Keras implementation uses a flatten layer).

Our design adds global pooling to decrease the output size of the networks (this is in line with most ConvNets but contrasts with VGGs, in which flatten layers are utilized extensively). Notably, we also insert an RNN module to evaluate of our proposed approach. The module comprises a reshape layer and an RNN layer as described in subsection \ref{rnn}. Moreover, we add a Gaussian noise layer to increase sample variation and prevent overfitting. In the fully connected layer, the number of neurons varies , (e.g., 256, 512, 1024 or 2048); the actual value is based on the specific experiment. The softmax layer uses the same number of outputs as the number of iNaturalist'19 categories.

All the networks' layers from the ConvNets are unfrozen; we reuse only the models' architectures and pretrained weights (these ConvNets are pretrained on the ImageNet dataset~\cite{deng2009imagenet,russakovsky2015imagenet}). Reusing the trained weights offers several advantages because retraining from scratch takes days, weeks or even months on a large dataset such as ImageNet. Typically, transfer learning can be used for most applications based on the concept that the early layers act like edge and curve filters; once trained, ConvNets can be reused on other, similar datasets~\cite{zeiler2014visualizing}. However, when the target dataset differs substantially from the pretrained dataset, retraining or fine-tuning can increase the accuracy. To distinguish these ConvNets from the original ones, we refer to each model as a fine-tuned model.

\begin{figure}[!ht]
\begin{center}
\includegraphics[width=0.5\textwidth]{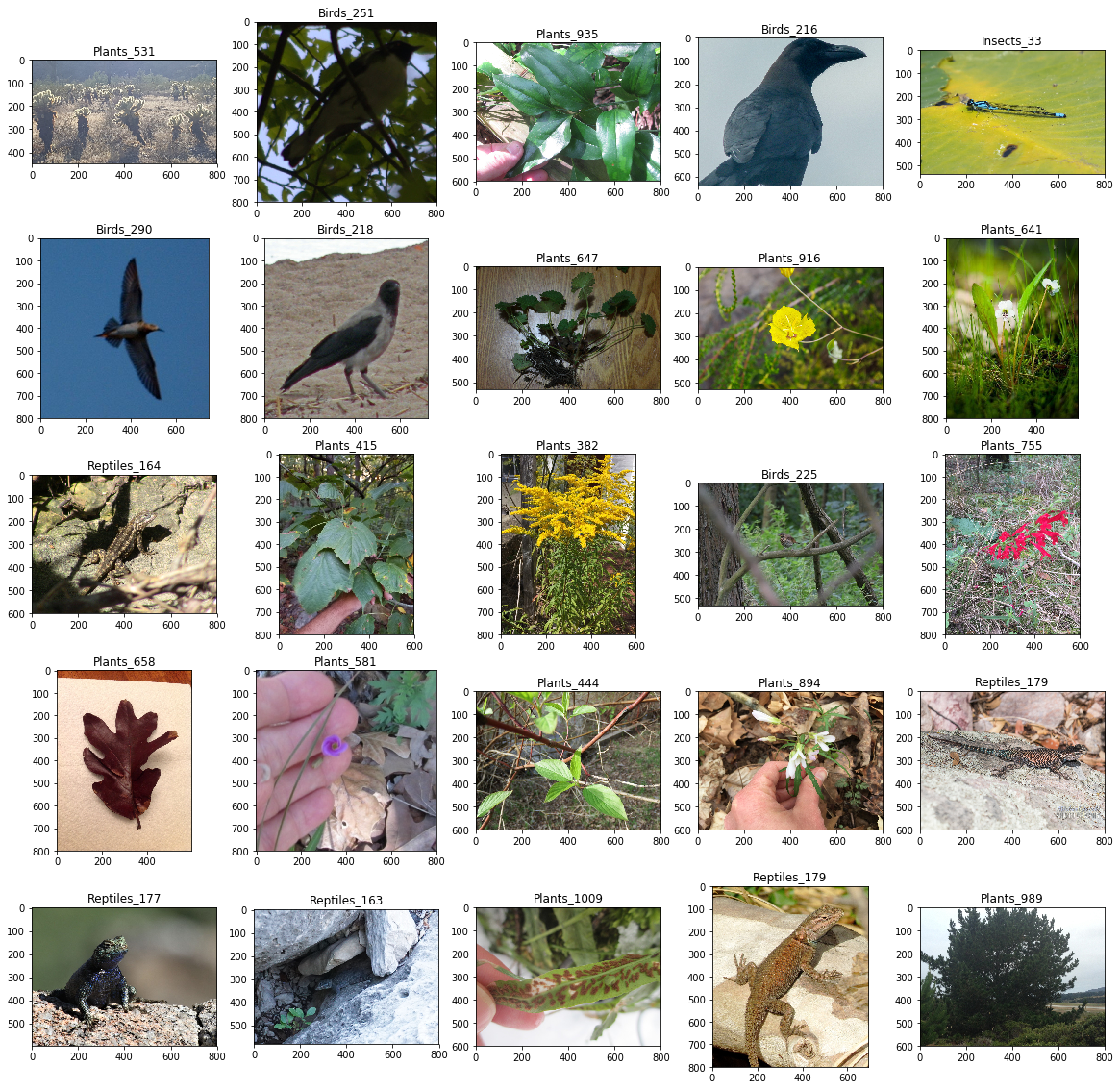}
\caption{Random Samples from iNaturalist'19 dataset. Each image is labeled with the species name and its subcategory}
\label{fig:iNat19_dataset}
\end{center}
\end{figure}

We conduct our experiments on the iNaturalist'19 dataset, which was originally compiled for the iNaturalist Challenge, conducted at the 6th Fine-grained Visual Categorization (FGVC6) workshop at CVPR 2019. In the computer vision area, FGVC has been a focus of researchers since approximately 2011~\cite{wah2011caltech,yao2011combining,khosla2011novel}, although research on similar topics appeared long before~\cite{collin2005subordinate,gillebert2009subordinate}. FGVC or subordinate categorization aims to classify visual objects at a more subtle level of detail than basic level categories~\cite{hillel2007subordinate}, for example, bird species rather than just birds~\cite{wah2011caltech}, dog types~\cite{khosla2011novel}, car brands ~\cite{KrauseStark3DRR2013}, and aircraft models~\cite{maji13finegrained}. The iNaturalist dataset was created in line with the development of FGVC~\cite{van2018inaturalist}, and the dataset is comparable with ImageNet regarding size and category variation. The specific dataset used in this research (iNaturalist'19) focuses on more similar categories than previous versions and is composed of 1,010 species collected from approximately two hundred thousand real plants and animals. Figure~\ref{fig:iNat19_dataset} shows some random images from this dataset and indicates the species shown as well as their respective classes and subcategories.

The dataset is randomly split into training and test sets at a ratio of $80:20$. In addition, the images are resized to appropriate resolutions, e.g., for the InceptionV3 standard, the rescaled images have a size of $299 \times 299$. The resolution is also increased to $401 \times 401$ or even $421 \times 421$. In the Gaussian noise layer, we set the amount of noise to 0.1, and in the fully connected layer, we chose $1,024$ neurons based on experience, because it is impractical to evaluate all the layers with every setting.

We configured a Jupyter Notebook server running on a Linux operating system (OS) using 4 GPUs (GeForce\textsuperscript{\textregistered} GTX 1080 Ti) each with 12 GB of RAM. For coding, we used Keras with a TensorFlow backend~\cite{abadi2016tensorflow} as our platform. Keras is written in the Python programming language and has been developed as an independent wrapper that runs on top of several backend platforms, including TensorFlow. The project was recently acquired by Google Inc. and has become a part of TensorFlow.

\begin{figure}[ht!]
\begin{center}
\includegraphics[width=0.5\textwidth]{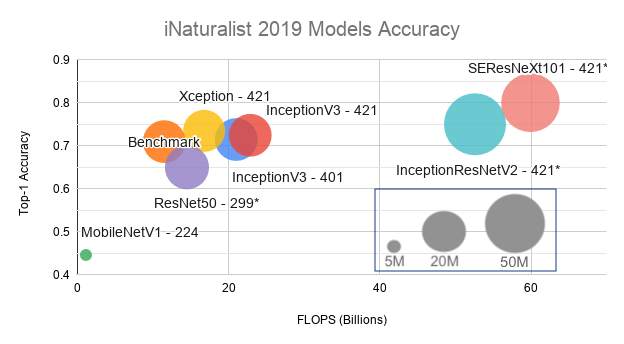}
\caption{iNaturalist'19 Model Accuracy. Here, ``Benchmark`` denotes the InceptionV3 result using the default input setting (an image resolution of $299 \times 299$). The sizes of the input images for the other models are indicated along with their respective names. For example, Xception-421 indicates that the input images for that model have been rescaled to $421 \times 421$.}
\label{fig:iNaturalist19}
\end{center}
\end{figure}

Figure~\ref{fig:iNaturalist19} shows the results of this experiment in which the top-1 accuracy is plotted against the floating-point operations per second (FLOPS). The size of each model or the total number of parameters is also displayed. The top-1 accuracy was obtained by submitting predictions to the challenge website, obtaining the top-1 error level from the private leaderboard and subtracting the result from 1. During a tournament, the public leaderboard is computed based on 51\% of the official test data. After the tournament, the private leaderboard contains a summary of all the data.

As expected, a higher image resolution yields greater accuracy but also uses more computing power for the same model (InceptionV3). As a side note, our benchmark achieved an accuracy of $0.7097$, which is a marginal gap from the benchmark model on the Challenge website ($0.7139$). Because we have no knowledge of the organizers' ConvNet designs, settings and working environments, we cannot delve further into the reasons for this difference. Later, we increased the image resolutions from $299 \times 299$ to $401 \times 401$ and $421 \times 421$ and switched the fine-tuned models. Using Xception-421, our model obtained an accuracy of approximately $0.7347$.

Because our server is shared, training takes approximately one week each time. This is the reason why we could increase the image size to only $421 \times 421$. In addition, we did not obtain the results for SEResNeXt101-421, Inception-ResNetV2-421 and ResNet50-299in this experiment, but projected their approximated accuracies; we will use these models in later experiments. Additionally, because adding an RNN module would substantially increase the training time, the RNN module is not analyzed on the iNaturalist'19 dataset.

\subsection{Experiment 2}
\label{experiments2}

As mentioned in the previous section, most of the research regarding RNNs has focused on sequential data or time series. Despite the little attention paid to using RNNs with images, the main goal is to generate sequences of pixels rather than direct image recognition. Our approach differs significantly from typical image recognition models, in which all the image pixels are presented simultaneously rather than in several time steps.

We systematically evaluated the proposed design that utilizes distinct recurrent neural networks. These models include a typical RNN, an advanced GRU and a bidirectional RNN--BiLSTM---and we compared them against a standard (STD) model without an RNN module. In addition, we selected representative fine-tuned models, namely, InceptionV3, Xception, ResNet50, Inception-ResNetV2, MobileNetV1, VGG19 and SEResNeXt101, for comparison and analysis.

\begin{figure}[hbt!]
\begin{center}
\includegraphics[width=0.5\textwidth]{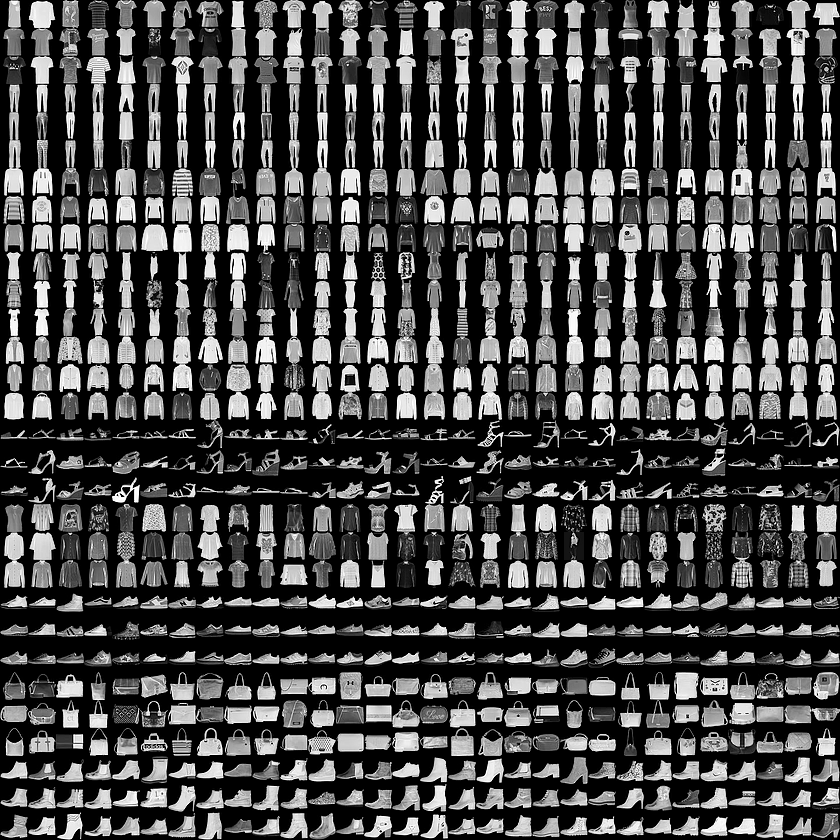}
\caption{Some Samples from the Fashion-MNIST dataset}
\label{fig:fashion_mnist_sprite}
\end{center}
\end{figure}

In this experiment, we employ the Fashion-MNIST dataset, which was recently created by Zalando SE and intended to serve as a direct replacement for the MNIST dataset as a machine learning benchmark because some models have achieved an almost perfect result of $100\%$ accuracy on MNIST. The Fashion-MNIST dataset contains the same amount of data as MNIST--$50,000$ training and $10,000$ testing samples--and includes 10 categories. Figure \ref{fig:fashion_mnist_sprite} visualizes how this dataset looks; each sample is a $28 \times 28$ grayscale image.

Our initial design (as discussed previously) is reused with a highlighted note that the RNN module has been incorporated. The number of units for each RNN is set to 2,048 (in the BiLSTM, the number of units is 1,024); the time step number is simply one, which shows the entire image each time. In addition, because the image size in the Fashion-MNIST dataset is smaller than the desired resolutions (e.g., $244 \times 244$ or $299 \times 299$ for MobileNetV1 and InceptionV3), all the images are upsampled.

We performed these experiments on Google Colab\footnote{https://colab.research.google.com/, which began as an internal project built based on Jupyter Notebook but was opened for public use in 2018. At the time of this writing, the virtual environment supports only a single 12 GB NVIDIA Tesla K80 GPU.}, even though our server runs faster. The main reason is that Jupyter Notebook occupies all the GPUs for the first login section. In other words, only one program executes with full capability. In contrast, Colab can run multiple environments in parallel. A secondary reason is that the virtual environment allows rapid development (i.e., it supports the installation of additional libraries and can run programs instantly). All the experiments are set to run for 12 hours; some take less time before overfitting, but others take more than the maximum time allotted. We repeated each experiment 3 times. Moreover, in this study, we often submit the results to challenge websites and record only the highest accuracy rather than employing other metrics. The accuracy metric is defined as follows:

\begin{equation}
Accuracy=\frac{\mbox{Number of correct predictions}}{\mbox{Total numbers of predictions made}}.
\end{equation}

\begin{figure}[hbt!]
\begin{center}
\includegraphics[width=0.5\textwidth]{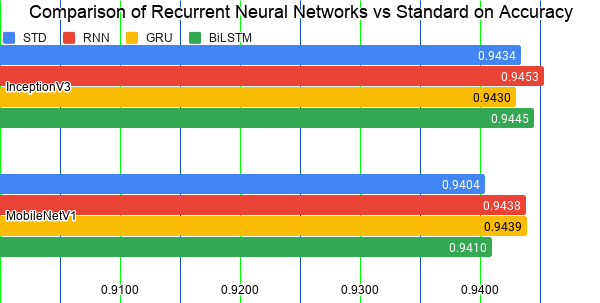}
\caption{Accuracy comparison of models integrated with recurrent neural networks vs. the standard model (STD). The integrated models significantly outperform the STD models.}
\label{fig:comparison_rnns_std}
\end{center}
\end{figure}

Table \ref{tab:rnns_inception_mobilenet} shows comparisons of the various models using different RNN modules built on InceptionV3 and MobileNetV1. The former is often chosen as a baseline model, whereas the latter is the lightest model used here in terms of parameters and computation. The results in Figure \ref{fig:comparison_rnns_std} show that the models with the additional RNN modules achieve substantially higher accuracy than do the STD models.

\begin{table}[]
\centering
\caption {\label{tab:rnns_inception_mobilenet}Accuracy comparison of models using distinct recurrent neural networks built on InceptionV3 and MobileNetV1}
\begin{tabular}{|l|c|c|c|c|}
\hline
                           & \multicolumn{4}{c|}{\textbf{Recurrent Neural Networks}} \\ \hline
\textbf{Fine-tuned Models} & STD          & RNN          & GRU         & BiLSTM      \\ \hline
InceptionV3               & 0.9434       & 0.9453       & 0.9430      & 0.9445      \\ \hline
MobileNetV1               & 0.9404       & 0.9438       & 0.9439      & 0.9410      \\ \hline
\end{tabular}
\end{table}

%

%
\begin{table}[]
\centering
\caption {\label{tab:rnn_std_vs_bilstm}Comparison of BiLSTM and STD on different fine-tuned models, including InceptionV3, Xception, ResNet50, Inception-ResNetV2, MobileNetV1, VGG19 and SEResNeXt101.}
\begin{tabular}{|l|c|c|c|c|}
\hline
\multirow{2}{*}{} & \multicolumn{4}{c|}{\textbf{Fine-tuned Models}}                                                                                                                                                                                           \\ \cline{2-5} 
                  & \textbf{\begin{tabular}[c]{@{}c@{}}Inception\\ V3\end{tabular}} & \textbf{Xception} & \textbf{\begin{tabular}[c]{@{}c@{}}ResNet\\ 50\end{tabular}}     & \textbf{\begin{tabular}[c]{@{}c@{}}InceptionResNet\\ V2\end{tabular}} \\ \hline
\textbf{STD}      & 0.9434                                                          & 0.9374            & 0.9375                                                           & 0.9364                                                                \\ \hline
\textbf{BiLSTM}   & 0.9445                                                          & 0.9383            & 0.9377                                                           & 0.9360                                                                \\ \hline
                  & \textbf{\begin{tabular}[c]{@{}c@{}}MobileNet\\ V1\end{tabular}} & \textbf{VGG19}    & \textbf{\begin{tabular}[c]{@{}c@{}}SEResNeXt\\ 101\end{tabular}} &                                                                       \\ \hline
\textbf{STD}      & 0.9404                                                          & 0.9075            & N/A                                                                &                                                                       \\ \hline
\textbf{BiLSTM}   & 0.9410                                                          & N/A                 & N/A                                                                &                                                                       \\ \hline
\end{tabular}
\end{table}

\begin{figure}[hbt!]
\begin{center}
\includegraphics[width=0.5\textwidth]{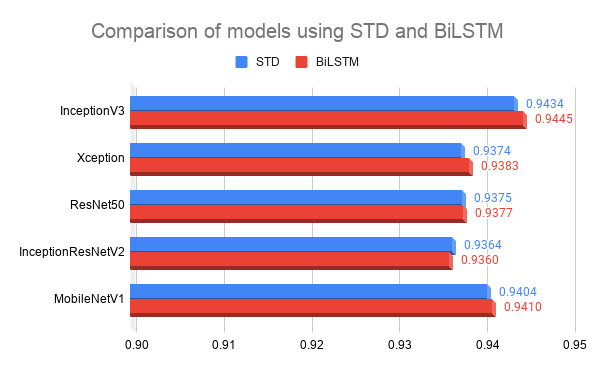}
\caption{Comparison of the accuracy of BiLSTM and STD models using InceptionV3, Xception, ResNet50, Inception-ResNetV2 and MobileNetV1. The BiLSTM models substantially outperform the STD models in several instances.}
\label{fig:comparison_std_bilstm}
\end{center}
\end{figure}

We also extended this experiment to include more models, including Xception, ResNet50, Inception-ResNetV2, VGG19 and SEResNeXt101. Table \ref{tab:rnn_std_vs_bilstm} shows the comparisons of models integrated with the BiLSTM module versus the standard models. Most of the models integrated with BiLSTMs significantly outperform the STDs, except for Inception-ResNetV2. Please note that training the VGG19 model required excessive training time; the model did not reach the same accuracy level as the other models even after 12 hours of training. Similarly, we were unable to obtain results for the VGG19-BiLSTM or SEResNeXt101 models. Therefore, Figure \ref{fig:comparison_std_bilstm} shows only the results for InceptionV3, Xception, ResNet50, Inception-ResNetV2 and MobileNetV1.

\subsection{Experiment 3}
\label{experiments3}
Often, when training a model, we randomly split a dataset into training and testing sets with a desired ratio. Then, we repeat our evaluation several times, and finally, obtain the results from one of the measurement methods, e.g., accuracy mean and standard deviation. However, in competitions such as those on Kaggle, a test set is completely separated from a training set. If we naively divide the training set into a training set and a validation set, we face a dilemma in which all the samples of the original training set cannot be used for training because a portion of the dataset is always needed for validation. To solve this problem, we apply k-fold validation to the training set by dividing the dataset into k subsets, of which one is reserved for validation. We expand this process and make predictions on the official test set. This method allows our models (one model for each set) to learn from all the images in the official training set.

We also attempted to increase model robustness via data augmentation techniques. The goal of such techniques is to transform an original training dataset into an expanded dataset whose true labels are known\cite{jackson2019style,zhong2020random,lopes2019improving}. Importantly, this teaches the model to be invariant to and uninfluenced by input variations \cite{lecun2012learning}. For example, flipping an image of a car horizontally does not change its corresponding category. We applied the augmentation approach from \cite{krizhevsky2012imagenet} on the test set. In this approach a sample is cropped multiple times and the model makes predictions for each instance. The procedure has recently become standard practice in image recognition tasks and is referred to as ``test time augmentation.`` In this study, we cropped the images at random locations instead of at only the four corners and the center. In addition, we utilized the most prevalent augmentation techniques for geometric and texture transformations, such as rotation, width/height shifts, shear and zoom, horizontal and vertical flips and channel shift.

\begin{figure}[htb!]
\begin{center}
\includegraphics[width=0.5\textwidth]{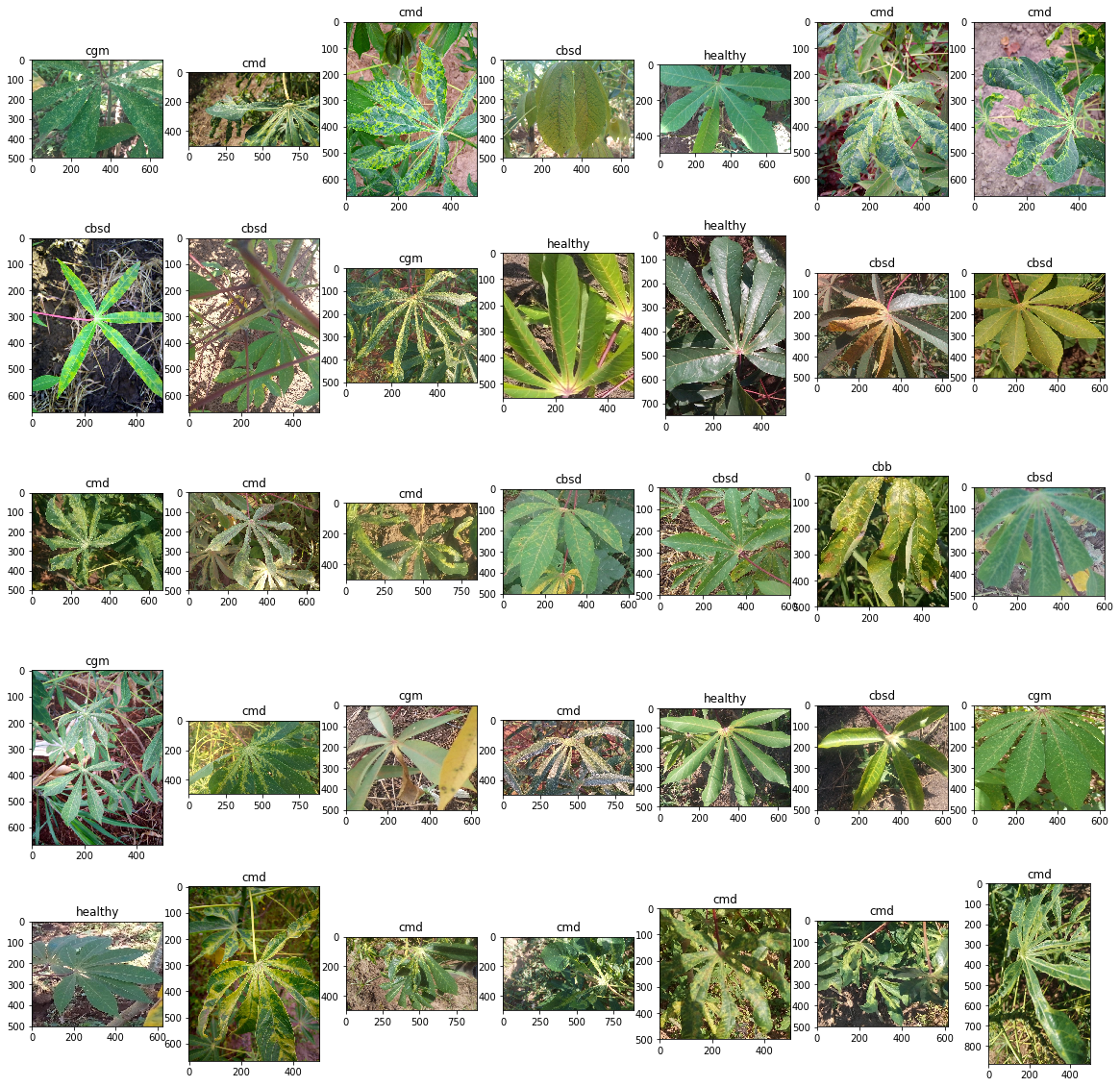}
\caption{Random Samples from the iCassava dataset. CMD, CGM, CBB and CBSD denote cassava mosaic disease, cassava green mite disease, cassava bacterial blight, and cassava brown streak disease, respectively.}
\label{fig:iCassava_dataset}
\end{center}
\end{figure}

We also applied state-of-the-art ensemble learning because the technique is generally more accurate than is prediction from a single model. We use a simple averaging approach to aggregate all the models (a.k.a AVG-3M). Additionally, it is important to ensure a variety of the fine-tuned models to increase the classifier diversity because combining multiple redundant classifiers would be meaningless. We finally chose the SERestNeXt101, Xception and Inception-ResNetV2 fine-tuned models, as these ConvNets yield higher results than do others. Please note that the RNN module (including the reshape and BiLSTM layers) is excluded.

One last crucial technique is our training strategy, which helps to reduce loss and increase accuracy by searching for a better global minimum (more details will be presented in the next section).

We evaluated our approach using the iCassava Challenge dataset, which was also compiled for the FGVC6 workshop at CVPR19. In the iCassava dataset, the leaf images of cassava plants are divided into 4 disease categories: cassava mosaic disease, cassava green mite disease, cassava bacterial blight, and cassava brown streak disease, and 1 category of healthy plants, comprising 9,436 labeled images. The challenge was organized on the Kaggle website \footnote{https://www.kaggle.com/c/cassava-disease}, ran from the 26th of April to the 2nd of June 2019, and attracted nearly 100 teams from around the world. The proposed models were evaluated on 3,774 official test samples, and the results were submitted to the website. The public leaderboard is computed from 40\% of the test data, whereas the private leaderboard is computed from all the test data. Figure \ref{fig:iCassava_dataset} shows some random samples from this dataset.

\begin{sidewaystable}
\centering
\caption {\label{tab:icassava_result}iCassava Result. The dataset is divided into 5 subsets. Four of the  subsets are combined to form a training set and the other subset is used as for validation. Each of these combinations (including training and validation sets) composes a dataset, named Sets 1--5. A ``Valid`` label denotes results obtained using valid data, while ``Private`` and ``Public`` represent results obtained from corresponding categories on the challenge website. Test--0 Crop indicates test results obtained without cropping, whereas Test--3 Crop indicates test results obtained using the 3 cropping methods. The best results for each subset are summarized in the table at the bottom left, including both the private and public results. The table on the bottom right represents the results obtained using AVG-3M for each individual set and when each result is combined with all the others
}
\tiny
\begin{tabular}{llllllllllll}
\hline
\multicolumn{1}{|l|}{\textbf{Models}}                                                                                        & \multicolumn{1}{l|}{}                       & \multicolumn{1}{l|}{\textbf{Set 1}}   & \multicolumn{1}{l|}{}                & \multicolumn{1}{l|}{\textbf{Set 2}}                       & \multicolumn{1}{l|}{}                              & \multicolumn{1}{l|}{\textbf{Set 3}}                 & \multicolumn{1}{l|}{}                & \multicolumn{1}{l|}{\textbf{Set 4}}   & \multicolumn{1}{l|}{}                                                            & \multicolumn{1}{l|}{\textbf{Set 5}}   & \multicolumn{1}{l|}{}                \\ \hline
\multicolumn{1}{|l|}{}                                                                                        & \multicolumn{1}{l|}{\textbf{Valid}}         & \multicolumn{1}{l|}{0.9043}           & \multicolumn{1}{l|}{}                & \multicolumn{1}{l|}{0.9105}                               & \multicolumn{1}{l|}{}                              & \multicolumn{1}{l|}{0.8963}                         & \multicolumn{1}{l|}{}                & \multicolumn{1}{l|}{0.9025}           & \multicolumn{1}{l|}{}                                                            & \multicolumn{1}{l|}{0.9061}           & \multicolumn{1}{l|}{}                \\ \cline{2-12} 
\multicolumn{1}{|l|}{}                                                                                        & \multicolumn{1}{l|}{}                       & \multicolumn{1}{l|}{\textbf{Private}} & \multicolumn{1}{l|}{\textbf{Public}} & \multicolumn{1}{l|}{\textbf{Private}}                     & \multicolumn{1}{l|}{\textbf{Public}}               & \multicolumn{1}{l|}{\textbf{Private}}               & \multicolumn{1}{l|}{\textbf{Public}} & \multicolumn{1}{l|}{\textbf{Private}} & \multicolumn{1}{l|}{\textbf{Public}}                                             & \multicolumn{1}{l|}{\textbf{Private}} & \multicolumn{1}{l|}{\textbf{Public}} \\ \cline{2-12} 
\multicolumn{1}{|l|}{}                                                                                        & \multicolumn{1}{l|}{\textbf{Test--0 Crop}} & \multicolumn{1}{l|}{0.9019}           & \multicolumn{1}{l|}{0.8874}          & \multicolumn{1}{l|}{0.9129}                               & \multicolumn{1}{l|}{0.8947}                        & \multicolumn{1}{l|}{0.9041}                         & \multicolumn{1}{l|}{0.8854}          & \multicolumn{1}{l|}{0.905}            & \multicolumn{1}{l|}{0.8814}                                                      & \multicolumn{1}{l|}{0.9041}           & \multicolumn{1}{l|}{0.8874}          \\ \cline{2-12} 
\multicolumn{1}{|l|}{\multirow{-4}{*}{\textbf{\begin{tabular}[c]{@{}l@{}}SEResNeXt\\ 101\end{tabular}}}}      & \multicolumn{1}{l|}{\textbf{Test--3 Crop}} & \multicolumn{1}{l|}{0.9023}           & \multicolumn{1}{l|}{0.8887}          & \multicolumn{1}{l|}{0.9143}                               & \multicolumn{1}{l|}{0.9026}                        & \multicolumn{1}{l|}{0.9015}                         & \multicolumn{1}{l|}{0.888}           & \multicolumn{1}{l|}{0.9032}           & \multicolumn{1}{l|}{0.89}                                                        & \multicolumn{1}{l|}{0.9121}           & \multicolumn{1}{l|}{0.896}           \\ \hline
\multicolumn{1}{|l|}{}                                                                                        & \multicolumn{1}{l|}{\textbf{Valid}}         & \multicolumn{1}{l|}{-}                & \multicolumn{1}{l|}{}                & \multicolumn{1}{l|}{-}                                    & \multicolumn{1}{l|}{}                              & \multicolumn{1}{l|}{0.9025}                         & \multicolumn{1}{l|}{}                & \multicolumn{1}{l|}{0.914}            & \multicolumn{1}{l|}{}                                                            & \multicolumn{1}{l|}{0.9158}           & \multicolumn{1}{l|}{}                \\ \cline{2-12} 
\multicolumn{1}{|l|}{}                                                                                        & \multicolumn{1}{l|}{}                       & \multicolumn{1}{l|}{\textbf{Private}} & \multicolumn{1}{l|}{\textbf{Public}} & \multicolumn{1}{l|}{\textbf{Private}}                     & \multicolumn{1}{l|}{\textbf{Public}}               & \multicolumn{1}{l|}{\textbf{Private}}               & \multicolumn{1}{l|}{\textbf{Public}} & \multicolumn{1}{l|}{\textbf{Private}} & \multicolumn{1}{l|}{\textbf{Public}}                                             & \multicolumn{1}{l|}{\textbf{Private}} & \multicolumn{1}{l|}{\textbf{Public}} \\ \cline{2-12} 
\multicolumn{1}{|l|}{}                                                                                        & \multicolumn{1}{l|}{\textbf{Test--0 Crop}} & \multicolumn{1}{l|}{-}                & \multicolumn{1}{l|}{-}               & \multicolumn{1}{l|}{-}                                    & \multicolumn{1}{l|}{-}                             & \multicolumn{1}{l|}{0.901}                          & \multicolumn{1}{l|}{0.89}            & \multicolumn{1}{l|}{0.9059}           & \multicolumn{1}{l|}{0.8986}                                                      & \multicolumn{1}{l|}{0.9081}           & \multicolumn{1}{l|}{0.9}             \\ \cline{2-12} 
\multicolumn{1}{|l|}{\multirow{-4}{*}{\textbf{Xception}}}                                                     & \multicolumn{1}{l|}{\textbf{Test--3 Crop}} & \multicolumn{1}{l|}{-}                & \multicolumn{1}{l|}{-}               & \multicolumn{1}{l|}{-}                                    & \multicolumn{1}{l|}{-}                             & \multicolumn{1}{l|}{0.9045}                         & \multicolumn{1}{l|}{0.8907}          & \multicolumn{1}{l|}{0.909}            & \multicolumn{1}{l|}{0.8966}                                                      & \multicolumn{1}{l|}{0.9045}           & \multicolumn{1}{l|}{0.8986}          \\ \hline
\multicolumn{1}{|l|}{}                                                                                        & \multicolumn{1}{l|}{\textbf{Valid}}         & \multicolumn{1}{l|}{-}                & \multicolumn{1}{l|}{}                & \multicolumn{1}{l|}{-}                                    & \multicolumn{1}{l|}{}                              & \multicolumn{1}{l|}{0.9043}                         & \multicolumn{1}{l|}{}                & \multicolumn{1}{l|}{}                 & \multicolumn{1}{l|}{0.9078}                                                      & \multicolumn{1}{l|}{0.9114}           & \multicolumn{1}{l|}{}                \\ \cline{2-12} 
\multicolumn{1}{|l|}{}                                                                                        & \multicolumn{1}{l|}{}                       & \multicolumn{1}{l|}{\textbf{Private}} & \multicolumn{1}{l|}{\textbf{Public}} & \multicolumn{1}{l|}{\textbf{Private}}                     & \multicolumn{1}{l|}{\textbf{Public}}               & \multicolumn{1}{l|}{\textbf{Private}}               & \multicolumn{1}{l|}{\textbf{Public}} & \multicolumn{1}{l|}{\textbf{Private}} & \multicolumn{1}{l|}{\textbf{Public}}                                             & \multicolumn{1}{l|}{\textbf{Private}} & \multicolumn{1}{l|}{\textbf{Public}} \\ \cline{2-12} 
\multicolumn{1}{|l|}{}                                                                                        & \multicolumn{1}{l|}{\textbf{Test--0 Crop}} & \multicolumn{1}{l|}{-}                & \multicolumn{1}{l|}{-}               & \multicolumn{1}{l|}{-}                                    & \multicolumn{1}{l|}{-}                             & \multicolumn{1}{l|}{0.9081}                         & \multicolumn{1}{l|}{0.8993}          & \multicolumn{1}{l|}{0.9098}                 & \multicolumn{1}{l|}{0.894}                                                       & \multicolumn{1}{l|}{0.9156}           & \multicolumn{1}{l|}{0.9026}          \\ \cline{2-12} 
\multicolumn{1}{|l|}{\multirow{-4}{*}{\textbf{\begin{tabular}[c]{@{}l@{}}InceptionResNet\\ V2\end{tabular}}}} & \multicolumn{1}{l|}{\textbf{Test--3 Crop}} & \multicolumn{1}{l|}{-}                & \multicolumn{1}{l|}{-}               & \multicolumn{1}{l|}{-}                                    & \multicolumn{1}{l|}{-}                             & \multicolumn{1}{l|}{0.8869}                         & \multicolumn{1}{l|}{0.8854}          & \multicolumn{1}{l|}{0.9094}                 & \multicolumn{1}{l|}{0.8962}                                                      & \multicolumn{1}{l|}{0.9151}           & \multicolumn{1}{l|}{0.9026}          \\ \hline
                                                                                                              &                                             &                                       &                                     &                                                           & \textbf{Private}                                                   & \textbf{Public}                                                    &                                      &                                       &                                                                                  & \textbf{Private}                                      & \textbf{Public}                                     \\ \cline{5-7} \cline{10-12} 
                                                                                                              &                                             &                                       & \multicolumn{1}{l|}{}                & \multicolumn{1}{l|}{\textbf{Set 5}}                       & \multicolumn{1}{l|}{0.9121}                        & \multicolumn{1}{l|}{0.896}                          &                                      & \multicolumn{1}{l|}{}                 & \multicolumn{1}{l|}{\textbf{\begin{tabular}[c]{@{}l@{}}AVG-3M\\ SET 5\end{tabular}}} & \multicolumn{1}{l|}{0.9240}           & \multicolumn{1}{l|}{0.9125}          \\ \cline{5-7} \cline{10-12} 
                                                                                                              &                                             &                                       & \multicolumn{1}{l|}{}                & \multicolumn{1}{l|}{\textbf{Set 4}}                       & \multicolumn{1}{l|}{0.905}                         & \multicolumn{1}{l|}{0.8814}                         &                                      & \multicolumn{1}{l|}{}                 & \multicolumn{1}{l|}{\textbf{All}}                                                & \multicolumn{1}{l|}{0.9324}           & \multicolumn{1}{l|}{0.9165}          \\ \cline{5-7} \cline{10-12} 
                                                                                                              &                                             &                                       & \multicolumn{1}{l|}{}                & \multicolumn{1}{l|}{\textbf{Set 3}}                       & \multicolumn{1}{l|}{0.9041}                        & \multicolumn{1}{l|}{0.8854}                         &                                      & \multicolumn{1}{l|}{}                 & \multicolumn{1}{l|}{\textbf{\begin{tabular}[c]{@{}l@{}}AVG-3M\\ SET 4\end{tabular}}} & \multicolumn{1}{l|}{0.9235}           & \multicolumn{1}{l|}{0.9099}          \\ \cline{5-7} \cline{10-12} 
                                                                                                              &                                             &                                       & \multicolumn{1}{l|}{}                & \multicolumn{1}{l|}{\textbf{Set 2}}                       & \multicolumn{1}{l|}{0.9143}                        & \multicolumn{1}{l|}{0.9026}                         &                                      & \multicolumn{1}{l|}{}                 & \multicolumn{1}{l|}{\textbf{All}}                                                & \multicolumn{1}{l|}{\textbf{0.9368}}  & \multicolumn{1}{l|}{0.9198}          \\ \cline{5-7} \cline{10-12} 
                                                                                                              &                                             &                                       & \multicolumn{1}{l|}{}                & \multicolumn{1}{l|}{\textbf{Set 1}}                       & \multicolumn{1}{l|}{0.9023}                        & \multicolumn{1}{l|}{0.8887}                         &                                      & \multicolumn{1}{l|}{}                 & \multicolumn{1}{l|}{\textbf{\begin{tabular}[c]{@{}l@{}}AVG-3M\\ SET 3\end{tabular}}} & \multicolumn{1}{l|}{0.9121}           & \multicolumn{1}{l|}{0.9079}          \\ \cline{5-7} \cline{10-12} 
                                                                                                              &                                             &                                       & \multicolumn{1}{l|}{}                & \multicolumn{1}{l|}{\cellcolor[HTML]{C0C0C0}\textbf{All}} & \multicolumn{1}{l|}{\cellcolor[HTML]{C0C0C0}0.928} & \multicolumn{1}{l|}{\cellcolor[HTML]{C0C0C0}0.9139} &                                      & \multicolumn{1}{l|}{}                 & \multicolumn{1}{l|}{\textbf{All}}                                                & \multicolumn{1}{l|}{0.9341}           & \multicolumn{1}{l|}{0.9258}          \\ \cline{5-7} \cline{10-12} 
\end{tabular}
\end{sidewaystable}

The iCassava official training set is split into 5 subsets for k-fold cross validation, in which one subset is reserved for testing and the others are used for training in turn. We performed these experiments on a server using a GeForce\textsuperscript{\textregistered} GTX 1080 Ti graphics card with 4 GPUs, each with 12 GB of RAM.

When evaluating the test set, we upsampled the images to a higher resolution ($540 \times 540$) and then randomly cropped them to the input size ($501 \times 501$). We varied the numbers of crops among 1, 3, 5, 7 and 9, and finally selected just 3 crops, as this option yielded higher accuracy than did the other cropping choices in most subsets. We collected the results for two methods, one without cropping and the other with 3 crops, and selected the one with the higher accuracy for each set. Then, we averaged the results for all the sets to provide a result for the entire dataset. The overall result achieved an accuracy of $0.928$.

\begin{figure}[hbt!]
\begin{center}
\includegraphics[width=0.5\textwidth]{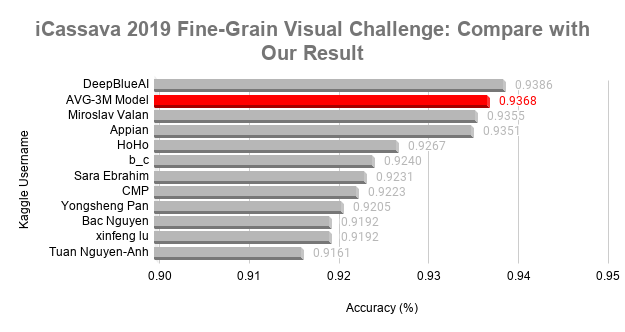}
\caption{Comparison of our AVG-3M result on the iCassava challenge dataset with those of the top-10 teams}
\label{fig:icassava_challenge}
\end{center}
\end{figure}

Furthermore, we applied the AVG-3M design to sets number 3, 4 and 5 and averaged each of these results with all other sets. Table \ref{tab:icassava_result} shows the results for this experiment. As the table shows, employing our AVG-3M model for these sets improves the accuracy. The combination of AVG-3M on Set 4 with the remaining sets yields the highest result ($0.9368$). It is critical to note that even though the public leaderboard result for all sets with the AVG-3M of Set 3 yields a higher result than that of Set 4, eventually, the latter achieved a better result. Therefore, choosing the submission for the final evaluation is efficient and essential. Figure~\ref{fig:icassava_challenge} shows our results in comparison with those of other top-10 teams.

\subsection{Experiment 4}
\label{experiments4}
\begin{figure}[htb!]
\begin{center}
\includegraphics[width=0.5\textwidth]{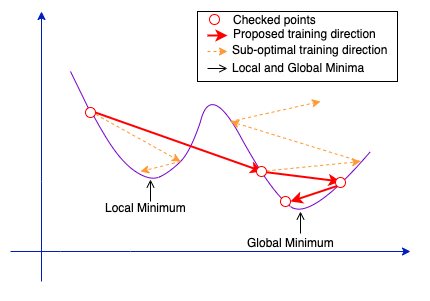}
\caption{Training Strategy. We start with a moderate learning rate so that training does not become stuck in a local minimum. Then, we reduce the learning rate for the 2nd and 3rd iterations. If the learning rate is too large, the global minimum might be ignored; however, if it is too small, the training time becomes excessive.}
\label{fig:optimals}
\end{center}
\end{figure}

In this subsection, we address the overall improvement of our model using a learning rate strategy. We apply the Adam optimizer~\cite{kingma2014adam}, which is an advanced optimizer in the deep learning area. The computations are as follows:
\begin{gather}
w_t=w_{t-1}-\eta_t\cdot\frac{m_t}{(\sqrt{v_t}+\hat{\epsilon})},\\
\eta_t=\eta\cdot\frac{\sqrt{1-\beta_2^t}}{1-\beta_1^t},\\
m_t=\beta_1\cdot m_{t-1}+(1-\beta_1)\cdot g_t,\\
v_t=\beta_2\cdot v_{t-1}+(1-\beta_2)\cdot g_t^2,
\end{gather}
where $w$ and $\eta$ are the weight and the learning rate of the neural networks, respectively; $m$, $v$ and $g$ are the moving averages and gradient of the current mini-batch; and the betas ($\beta_1$, $\beta_2$) and epsilon $\epsilon$ are set to $0.9$, $0.999$ and $10^{-8}$, respectively.

We used Keras to implement the models; on that platform, the formula for computing the learning rate with decay is

\begin{equation}
\eta=\eta\cdot\frac{1}{1+decay\cdot iterations}. 
\end{equation}
\begin{figure}[htb!]
\begin{center}
\includegraphics[width=0.5\textwidth]{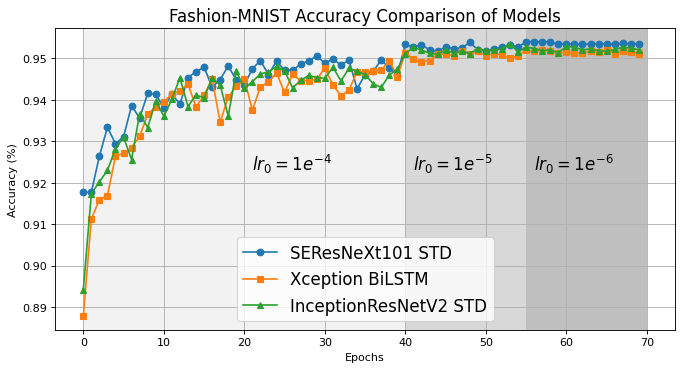}
\caption{ Model Accuracy on the Fashion-MNIST dataset Using Our Training Strategy. The models are trained with a learning rate starting at $1e^{-4}$ for 40 epochs. The RNN modules (STD, RNN, LSTM and GRU) are compared. The best module is selected for each model, e.g., the BiLSTM for Xception. The models are reloaded with the highest check points and trained two more times for 15 epochs each at learning rates of $1e^{-5}$ and $1e^{-6}$, respectively.}
\label{fig:fa_mnist_3models}
\end{center}
\end{figure}

Choosing an appropriate learning rate is both essential and critical, because the training time is often dramatically reduced with an appropriate learning rate. However, selecting an appropriate learning rate is difficult. When the step size is too large, the global minimum might be ignored, whereas a too-small step size can result in an excessive training time. In these experiments, we started with a moderate learning rate and trained the model until the accuracy stopped improving. Then, we reduced the learning rate, reloaded the ConvNet weights that achieved the highest accuracy and repeated this process for two more times. Figure~\ref{fig:optimals} illustrates our proposed training procedure.

We evaluated these experiments on the Cifar-10 and Fashion-MNIST datasets. Initially, the learning rate was set to $1e^{-4}$ and then sequentially changed to $1e^{-5}$ and $1e^{-6}$. The decay was derived by dividing the learning rate by the number of epochs.

Figure~\ref{fig:fa_mnist_3models} shows the performances of the top 3 models on Fashion-MNIST using SEResNeXt101 STD, Xception LSTM and Inception-ResNetV2 STD and the figure excludes the results of other ConvNets discussed in the previous sections. The three RNN variations were also analyzed against the STD models, and only the model with the highest accuracy is shown in the figure. The accuracies are effectively increased after the transition. SEResNeXt101 achieved a result of $0.9541$ under the STD setting.

This result was further improved to $0.9585$ when using the E2E-3M model The design is shown in Figure \ref{fig:mul_models}, and the steps are detailed in Algorithm \ref{alg:e2e3m}. The settings were as follows: the fully connected layer (after the concatenation layer) had 4,096 neurons; the LeakyReLU had a slope of 0.2 and dropout was set at 0.5. Please note that when reloading weights, we sometimes need to convert the weights to Pickle format rather than Keras' standard HDF5 format because the networks' weights are too large.

\begin{algorithm}[!htb]
\caption{E2E-3M}
\label{alg:e2e3m}
\begin{algorithmic}[1]
\Require
{A training set with c categories $D:=(a_1,b_1), (a_2,b_2)\dots(a_n,b_n)$
} \;
\Ensure{Class Predictions}
\State Step 1: Train the Level-1 classifiers
\StateX Number of L1 learners = m
\State Step 2: Train m fine-tuned models
\State Step 3: Select top-3 models
\State Step 4: Reload the weights of the three models
\State Step 5: Construct a new dataset of predictions
\For{i = 1 to $n$}
\begin{gather*}
M_i=(a_i^{'},b_i)
\end{gather*}
\StateX where:
\begin{gather*}
a_i^{'}=[\Delta_1(a_i)\ \Delta_2(a_i)\ \Delta_3(a_i)]^T\\
\Delta_j=[\delta_1\ \delta_2 \dots \delta_c]
\end{gather*}
\EndFor
\State Step 6: Train $\lambda$ neurons in the fully connected neural networks $(M)$
\begin{equation*}
    \mu_j = \sum\limits_{i=0}^{3c-1} \delta_{i} w_{ij},\ 1 \leq j \leq \lambda
    \end{equation*}
\State Step 7: Apply LeakyReLU activation (slope = $k$)
\begin{equation*}
        \eta_j =
        \begin{cases}
          \mu_j, & \text{if}\ \mu_j \geq 0 \\
          k\mu_j, & \text{otherwise}
        \end{cases}
      \end{equation*}
\State Step 8: Regularize using dropout (rate = $p$)
\begin{gather*}
\upsilon_j = \gamma \eta_j
\end{gather*}
\StateX where $\gamma$ is a gating variable 0-1 that follows a Bernoulli distribution with $P(\gamma=1) = p$
\State Step 9: Train the Final Fully Connected Neural Networks
\State Step 10: Apply Softmax
\end{algorithmic}
\end{algorithm}
\begin{figure}[hbt!]
\begin{center}
\includegraphics[width=0.5\textwidth]{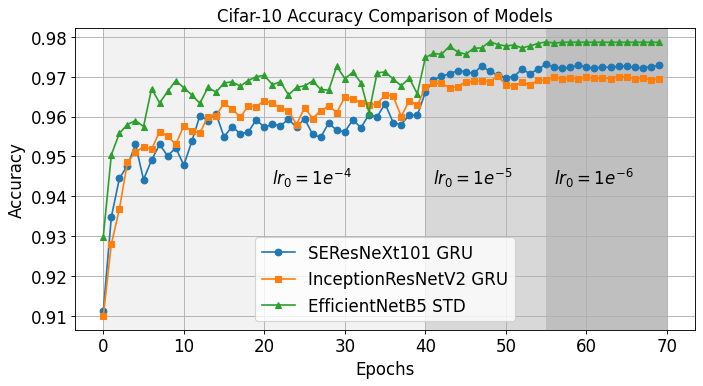}
\caption{Model Accuracies on Cifar-10 Using Our Training Strategy. The models are trained with a learning rate starting at $1e^{-4}$ for 40 epochs. Then the RNN modules (including STD, RNN, LSTM and GRU) are compared. The best module is selected for each model, (e.g., STD in the case of EfficientNetB5). The models are then reloaded with the highest checkpoints and trained twice more for 15 epochs each at learning rates of $1e^{-5}$ and $1e^{-6}$, respectively.}
\label{fig:cifar10_3models}
\end{center}
\end{figure}

We conducted experiments in the same manner on the Cifar-10 dataset. The results are shown in Figure~\ref{fig:cifar10_3models}. Please note that we also added EfficientNetB5, as this ConvNet is one of the latest models in the field. Using EfficientNetB5, we achieved an accuracy of $0.9788$ under the STD setting.

In summary, our learning rate strategy differs from conventional methods, where the learning rates are predefined based on experiments. Applying an adaptive learning rates offers more flexibility to adjust the learning rates automatically during training. Our approach applies the Adam optimizer but in an intuitive way. After training for some epochs with the optimizer, the models begin overfitting. Thus, in our method, we reduce the learning rate and reload the best model, which leverages the accuracy to a higher point.

\subsection{Experiment 5}
\label{experiments5}
In this subsection, our setup includes a variation of the softmax layer in which only the outputs of the most active neurons are used for prediction. We observe that in multicategory prediction, often, only a few or even just one confidence value on a category is large enough to be meaningful, while others are very small, i.e., a confidence level of nearly zero percent. For this reason, we propose eliminating meaningless confidence levels by zeroing them before use in ensembling. We compared this approach (namely, EXT-Softmax) with a typical method where multiple predictions are averaged to obtain a final prediction (AVG-Softmax). The essential steps are illustrated in Algorithm \ref{alg:pruning}.

\begin{algorithm}[!htb]
\caption{Pruning}
\label{alg:pruning}
\begin{algorithmic}[1]
\Require{Prediction Confidence $A$}
\Ensure{Pruned $A$}
\Statex
\Function{Prune}{$A$}
\State {$N$ $\gets$ {$length(A)$}}
\State {$M$ $\gets$ {$length(A[0])$}}
\For{$i \gets 0$ to $N-1$}
\State {$indexMax$ $\gets$ {$findIndexOfMaxValue(A)$}}
\For{$j \gets 0$ to $M-1$}
\If {$A[i] \neq indexMax$}
\State $A[i]\gets 0$
\EndIf
\EndFor
\EndFor
\State \Return {$A$}
\EndFunction
\end{algorithmic}
\end{algorithm}

The results on Fashion-MNIST are $0.9592$ and $0.9591$ without improvement using the proposed approach. However, on Cifar-10, the accuracy is higher, ranging from $0.9833$ to $0.9836$. Tables \ref{tab:fashion_mnist_list} and \ref{tab:cifar10_list} show the latest achievements on these datasets. Please note that the Fashion-MNIST results were voluntarily submitted and were not officially verified. However, we analyzed each profile and selected only results that are supported by publications. Importantly, the dataset was changed recently 
to eliminate duplicates
; our results might have been even higher if the previous version had been used.

\begin{table}[]
\centering
\caption {\label{tab:fashion_mnist_list}List of latest achievements on Fashion-MNIST. The results were submitted to the official website for the Fashion-MNIST dataset. ``Classifier`` indicates the main method that was used to achieve the result.}

\begin{tabular}{lccl}
\hline
\textbf{Classifier}              & \textbf{Accuracy} & \textbf{Submitter} \\\hline
WRN-28-10 + Random Erasing       & 0.963                 & @zhunzhong07       \\
WRN-28-10                        & 0.959                 & @zhunzhong07       \\
Dual path network with WRN-28-10 & 0.957                 & @Queequeg          \\
DENSER                           & 0.953                 & @fillassuncao      \\
MobileNet                        & 0.950                 & @Bojone            \\
CNN with optional shortcuts      & 0.947                 & @kennivich         \\
Google AutoML                    & 0.939                 & @Sebastian Heinz   \\
Capsule Network                  & 0.936                 & @XifengGuo         \\
VGG16                            & 0.935                 & @QuantumLiu        \\
LeNet                      & 0.934                 & @cmasch          \\
\hline 
\textbf{AVG-Softmax}             & \textbf{0.9592}       & N/A                  \\
\textbf{EXT-Softmax}             & \textbf{0.9591}       & N/A                  \\
\textbf{E2E-3M}                  & \textbf{0.9585}       & N/A          \\
\textbf{SeResNeXt101-STD}        & \textbf{0.9541}       & N/A          \\
\hline 
\end{tabular}
\end{table}

\begin{table}[]
\centering
\caption {\label{tab:cifar10_list}List of recent achievements on Cifar-10 along with the results of our models. The proposed approach performs comparably to the top models.}
\begin{tabular}{lcc}
\hline
\textbf{Authors}                       & \textbf{Accuracy (\%)} \\\hline
Huang et al.~\cite{huang2019gpipe}    & 99.00                \\
Cubuk et al.~\cite{cubuk2019autoaugment}    & 98.52                \\
Nayman et al.~\cite{nayman2019xnas}      & 98.40                \\
\textbf{EXT-Softmax}      & \textbf{98.36}                \\
\textbf{AVG-Softmax}      & \textbf{98.33}                \\
\textbf{EfficientNetB5-STD}   & \textbf{97.88}                \\
Yamada et al.~\cite{yamada2018shakedrop} & 97.69                \\
DeVries et al.~\cite{devries2017improved} & 97.44                \\
\textbf{SEResNeXt101-GRU}   & \textbf{97.31}                \\
\textbf{Inception-ResNetV2-GRU}   & \textbf{97.02}                \\
Zhong et al.~\cite{zhong2017random}       & 96.92                \\
Liang et al.~\cite{liang2018drop}     & 96.55                \\
Huang et al.~\cite{huang2017densely}        & 96.54                \\
Graham~\cite{graham2014fractional}         & 96.53                \\
Zhang et al.~\cite{zhang2017residual}         & 96.23               \\\hline
\end{tabular}
\end{table}

\subsection{Experiment 6}
\label{extension}
In this subsection, we extend our experiments from an ensemble with only the three most accurate ConvNet models to a broader number of ConvNet models because a greater variety of predictions may yield better results. We also evaluated our models to the following more challenge datasets, i.e., the Street View House Numbers (SVHN) \cite{netzer2011reading} and Cifar-100 \cite{krizhevsky2009learning} datasets, in addition to Cifar-10. SVHN offers two versions: an original version with roughly similar number of samples as Cifar-10 and an extra version that increases the number of samples by an order of magnitude. Cifar-100, as the name suggests, expands the number of categories from ten to one hundred.

We use the same settings for Cifar-10 and Cifar-100 and slightly adjusted these configurations for SVHN. In addition to eliminating the flipping augmentations (since the numbers are changed even become another number e.g. number 2 looks similar to number 5 after flipping), we started the learning rate at $1e^{-3}$ instead of $1e^4$. Moreover, because the required training time is 10 times longer, we repeated the process only once rather than twice. For the specific settings, please refer to the provided source code.

Table \ref{tab:extend} compares the performances of our models with those of other approaches on Cifar-10, Cifar-100 and SVHN. For Cifar-10, our ensemble outperforms the previous state-of-the-art architecture \cite{nayman2019xnas} by a large margin. Across all the datasets, the neural architecture search seems to work well on small datasets but appears to struggle on larger datasets as search space grows exponentially. For Cifar-100, our method is much better than Autoaugment \cite{cubuk2019autoaugment}, which requires approximately 5000 hours of training for image augmentation (we trained each model for few days to one week that depends on the model size). In addition, we attempted to evaluate our model on SVHN. To match recent works, we utilized the extra dataset. The results of our first attempt were as good as the population-based augmentation method\cite{ho2019population}. For our second attempt, we upgraded our server to the latest libraries, which improved the accuracy to match that of Randaugment \cite{cubuk2020randaugment}. Additionally, our method outperforms the abovementioned methods on Cifar-10.

\begin{table*}[t]
\centering
\caption {Results on Cifar-10, Cifar-100 and SVHN}
\label{tab:extend}
\begin{tabular}{lclclcl} 
\hline
\multicolumn{7}{|c|}{Datasets}                                                                                                                                                                                                                                                                                       \\ 
\hline
\multicolumn{2}{|c|}{Cifar-10}                                                              & \multicolumn{2}{c|}{Cifar-100}                                                                       & \multicolumn{3}{c|}{SVHN}                                                                                         \\ 
\hline
\multicolumn{1}{|l|}{Method}                        & \multicolumn{1}{c|}{Accuracy}        & \multicolumn{1}{l|}{Method}                                  & \multicolumn{1}{c|}{Accuracy}        & \multicolumn{1}{l|}{Method}                        & \multicolumn{2}{c|}{Accuracy}                                \\ 
\hline
\multicolumn{1}{|l|}{Randaugment~\cite{cubuk2020randaugment}}                   & \multicolumn{1}{c|}{0.9850}           & \multicolumn{1}{l|}{~\cite{cubuk2020randaugment}}                             & \multicolumn{1}{c|}{-}               & \multicolumn{1}{l|}{~\cite{cubuk2020randaugment}}                   & \multicolumn{2}{c|}{0.9900}                                    \\
\multicolumn{1}{|l|}{Population based augmentation~\cite{ho2019population}} & \multicolumn{1}{c|}{-}               & \multicolumn{1}{l|}{~\cite{ho2019population}}           & \multicolumn{1}{c|}{-}               & \multicolumn{1}{l|}{~\cite{ho2019population}} & \multicolumn{2}{c|}{0.9899}                                  \\
\end{tabular}
\end{table*}

\subsection{Experiment 7}
\label{stateoftheart}
\begin{figure}[hbt!]
\begin{center}
\includegraphics[width=0.5\textwidth]{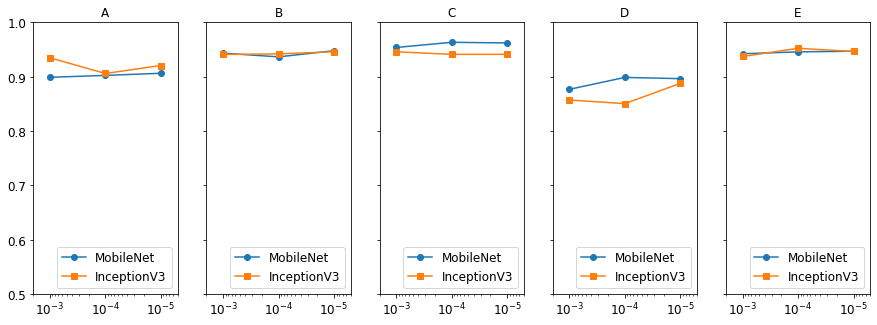}
\caption{Performances of MobileNet and InceptionV3 on leave-one-out cross validation using the five sets from Surrey}
\label{fig:surrey_lr}
\end{center}
\end{figure}

In this subsection, we compared our method with state-of-the-art approaches on the Surrey dataset \cite{pugeault2011spelling} using leave-one-out cross validation. The dataset contains more than $120,000$ color and depth images, each allocated into five subsets. The highest result of $0.9353$ accuracy was achieved by the recent approach in \cite{yang2020ddanet}. Our model outperforms the state-of-the-art model by $1.4\%$ when using only color images. Figure \ref{fig:surrey_lr} shows our results on the InceptionV3 and MobileNet fine-tuned ConvNets. In Table \ref{tab:surrey}, we compare the performance of our ensemble with those of other models.

Generally, these results demonstrate that our approach outperforms the state-of-the-art by large margins and is suitable for use in a variety of circumstances.
\begin{table*}[t]
\centering
\caption {Comparison of our method with previous state of the art approaches on Surrey dataset.}
\label{tab:surrey}
\begin{tabular}{ccccc}
                             &                         &                             &                                &                                \\ 
\hline
\multicolumn{1}{|c|}{Method} & \multicolumn{1}{c|}{F1} & \multicolumn{1}{c|}{Recall} & \multicolumn{1}{c|}{Precision} & \multicolumn{1}{c|}{Accuracy}  \\ 
\hline
\multicolumn{1}{|c|}{~\cite{pugeault2011spelling}}      & -                       & -                           & 0.4900                         & \multicolumn{1}{c|}{-}         \\
\multicolumn{1}{|c|}{~\cite{keskin2012hand}}      & -                       & -                           & -                              & \multicolumn{1}{c|}{0.8430}    \\
\multicolumn{1}{|c|}{~\cite{kuznetsova2013real}}      & -                       & -                           & -                              & \multicolumn{1}{c|}{0.5700}    \\
\multicolumn{1}{|c|}{~\cite{dong2015american}}      & -                       & -                           & -                              & \multicolumn{1}{c|}{0.7330}    \\
\multicolumn{1}{|c|}{~\cite{zhang2013histogram}}      & -                       & -                           & -                              & \multicolumn{1}{c|}{0.5600}    \\
\multicolumn{1}{|c|}{~\cite{pedersoli2014xkin}}     & -                       & 0.7700                      & 0.7900                         & \multicolumn{1}{c|}{-}         \\
\end{tabular}
\end{table*}

\section{Conclusions}
\label{conclusions}
In this paper, we presented our core ideas for improving ConvNets by integrating RNNs as essential layers in ConvNets when designing end-to-end multiple-model ensembles that gain expertise from each individual ConvNet, an improved training strategy, and a softmax layer extension.

First, we propose integrating RNNs into ConvNets even though RNNs are mainly optimized for 1D sequential data rather than 2D images. Our results on the Fashion-MNIST dataset show that ConvNets with RNN, GRU and BiLSTM modules can outperform standard ConvNets. We employed a variety of fine-tuned models in a virtual environment, including InceptionV3, Xception, ResNet50, Inception-ResNetV2 and MobileNetV1. Similar results can be obtained using a dedicated server for the SEResNeXt101 and EfficientNetB5 models on the Cifar-10 and Fashion-MNIST datasets. Although adding RNN modules requires more computing power, there is a potential trade-off between accuracy and running time.

Second, we designed E2E-3M ConvNets that learn predictions from several models. We initially tested this model on the iNaturalist'19 dataset using only a single E2E-3M model. Then, we evaluated various models and image resolutions and compared them with the Inception benchmark. Then, we added RNNs to the model and analyzed the performances on the Fashion-MNIST dataset. Our E2E-3M model outperforms a standard single model by a large margin. The advantage of using an end-to-end design is that the model can run immediately in real time and it is suitable for system-on-a-chip platforms.

Third, we proposed a training strategy and pruning for the softmax layer that yields comparable accuracies on the Cifar-10 and Fashion-MNIST datasets.

Fourth, using the ensemble technique, our models perform competitively, matching previous state-of-the-art methods (accuracies of 0.99, 0.9027 and 0.9852 on SVHN, Cifar-100 and Cifar-10, respectively) even with limited resources.

Finally, our method outperforms other approaches on the Surrey dataset, achieving an accuracy of 0.949.

In the future, we plan to extend our models to include a greater variety of settings. For example, we will evaluate bidirectional RNN and bidirectional GRU modules. In addition, we are eager to test our models on other datasets e.g. ImageNet.

\ifCLASSOPTIONcompsoc
\section*{Acknowledgments}
\else
\section*{Acknowledgment}
\fi

This research is sponsored by FEDER funds through the program COMPETE (Programa Operacional Factores de Competitividade) and by national funds through FCT (Fundação para a Ciência e a Tecnologia) under the projects UIDB/00326/2020; The authors gratefully acknowledge the reviewers for their insightful comments.

\bibliographystyle{IEEEtran}
\bibliography{references}

\begin{IEEEbiography}
[{\includegraphics[width=1in,height=1.25in,clip,keepaspectratio]{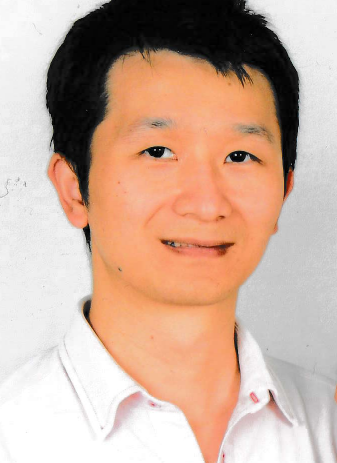}}]%
{Nguyen Huu Phong}
received the BSc in Physics from Vietnam National University, Hanoi and MSc degree in Information Technology from Shinawatra University, Thailand. He is currently a PhD student at the Department of Informatics Engineering of the University of Coimbra. His research interests include image recognition, action recognition, pattern recognition, machine learning and deep learning.
\end{IEEEbiography}

\begin{IEEEbiography}
[{\includegraphics[width=1in,height=1.25in,clip,keepaspectratio]{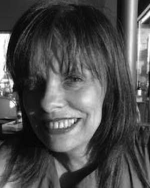}}]%
{Bernardete Ribeiro}
Bernardete Ribeiro (SM'15) received the Ph.D. and Habilitation degrees in informatics engineering from University of Coimbra. She is currently a Full Professor with University of Coimbra, also the Director of the Center of Informatics and Systems, University of Coimbra, also the former President of the Portuguese Association of Pattern Recognition, and also the Founder and the Director of the Laboratory of Artificial Neural Networks for over 20 years. Her research interests are in the areas of machine learning, and pattern recognition and their applications to a broad range of fields. She has been responsible/participated in several research projects both in international and national levels in a wide range of application areas. She is an IEEE SMC Senior Member, a member of the IARP International Association of Pattern Recognition, a member of the International Neural Network Society, a member of the APCA Portuguese Association of Automatic Control, a member of the Portuguese Association for Artificial Intelligence, a member of the American Association for the Advancement of Science, and a member of the Association for Computing Machinery. She received several awards and recognitions.
\end{IEEEbiography}

\end{document}